\documentclass[AMA,STIX1COL]{WileyNJD-v2}
\usepackage[utf8]{inputenc}
\usepackage{amsmath}
\usepackage{amssymb}
\usepackage{multirow}
\usepackage{PTusefulMacros}
\usepackage{stmaryrd}
\usepackage{multirow}
\usepackage{etoolbox}
\usepackage{hyperref}
\usepackage{subfig}
\usepackage[subfigure]{tocloft}

\usepackage{array}

\usepackage{easy-todo}

\usepackage{showframe}

\def\ptFiguresDirectory#1{./figures/#1}
\def\ivBrack#1{\left\llbracket{#1}\right\rrbracket} 
\newsavebox\CBox

\def\ptVecX#1#2#3{\left[\ptRowX{#1}{#2}{#3} \right]}

\def\ptRowX#1#2#3{{#1}_{#21},{#1}_{#22},\ldots,{#1}_{#2#3}}
\def\vec#1{#1}
\def\fdr{\mathrm{FDR}}
\def\fnr{\mathrm{FNR}}
\def\fOne{F_{1}}
\def\FWER{FWER}
\def\MIML{MIML}

 \newcounter{rowcntr}[table]
\renewcommand{\therowcntr}{\thetable.\arabic{rowcntr}}

\newcolumntype{N}{>{\refstepcounter{rowcntr}\therowcntr}c}

\AtBeginEnvironment{tabular}{\setcounter{rowcntr}{0}}

\articletype{ORIGINAL ARTICLE}%

\received{?}
\revised{?}
\accepted{?}
\begin{document}
\title{Bayes metaclassifier and Soft-confusion-matrix classifier in the task of multi-label classification.}

\author[1]{Pawel Trajdos*}

\author[2]{Marcin Majak}

\authormark{Trajdos P., Majak M.}

\address[1]{\orgdiv{Department of Systems and Computer Networks}, \orgname{Wroclaw University of Science and Technology}, \orgaddress{\state{Wroclaw}, \country{Poland}}}

\corres{*Corresponding author name \email{pawel.trajdos@pwr.edu.pl}}

\presentaddress{Wybrzeze Wyspianskiego 27, 50-370
 Wroclaw, Poland}

\abstract[Summary]{The aim of this paper was to compare soft confusion matrix approach and Bayes metaclassifier under the multi-label classification framework. Although the methods were successfully applied under the multi-label classification framework, they have not been compared directly thus far. Such comparison is of vital importance because both methods are quite similar as they are both based on the concept of randomized reference classifier. Since both algorithms were designed to deal with single-label problems, they are combined with the problem-transformation approach to multi-label classification. Present study included 29 benchmark datasets and four different base classifiers. The algorithms were compared in terms of 11 quality criteria and the results were subjected to statistical analysis.}

\keywords{multi-label classification, soft confusion matrix, Bayes metaclassifier}

\jnlcitation{\cname{%
\author{Trajdos Pawel}, 
\author{Majak Marcin}} (\cyear{2018}), 
\ctitle{Bayes metaclassifier and Soft-confusion-matrix classifier in the task of multi-label classification}, \cjournal{Computational Intelligence}, \cvol{?}.}

\maketitle

\section{Introduction}
Under the traditional supervised classification framework, the object is assigned to only one class. However, many real-world datasets contain objects that can be classified into several different categories. All these categories sum up to full description of the object and whenever one of them is missing, the information is incomplete. For example, the same image can be described using various tags, such as sea, beach and sunset. Such classification process is referred to as multi-label (ML) classification~\cite{Gibaja2014}. In the last 15 years, multi-label learning has found a~number of practical applications, including text classification~\cite{Jiang2012}, multimedia classification~\cite{Sanden2011} and bioinformatics~\cite{Wu2014}, to mention a~few.

Multi-label classification algorithms can be generally divided into two main groups, i.e. dataset transformation and algorithm adaptation approaches~\cite{Gibaja2014}.

The algorithm adaptation approach represents a~generalization of an existing multi-class algorithm, in which the generalized algorithm can be used to solve a~multi-label classification problem directly. The best known approaches from this group include multi-label nearest neighbors algorithm~\cite{Jiang2012}, ML Hoeffding trees~\cite{Read2012}, structured output support vector machines~\cite{Diez2014} and deep-learning-based algorithms~\cite{Wei2015}.

In the dataset transformation approach, a~multi-label problem is decomposed into a~set of single-label classification tasks. During the inference phase, the outputs of the underlying single-label classifiers are combined into a~multi-label prediction. One example of the dataset transformation approach is \textit{binary relevance} (BR) approach in which a~multi-label classification task is decomposed into a~set of \textit{one-vs-rest} binary classification problems~\cite{AlvaresCherman2010}. This algorithm is based on the assumption that labels are conditionally independent, which does not occur too often in the case of most real-life recognition problems. Despite this, the BR framework is still one of the most widespread multi-label classification methods~\cite{Tsoumakas2009}, due to its excellent scalability and acceptable classification quality. However, the method can be easily outperformed by algorithms that are adjusted for mutual relationships between labels~\cite{AlvaresCherman2010, Read2009}. An alternative technique to decompose a~multi-label classification task into a~set of binary classifiers is the \textit{label-pairwise} (LPW) scheme~\cite{Furnkranz2002}. In this approach, each pair of labels are assigned with a~\textit{one-vs-one} binary classifier. The outcome of the classifier is interpreted as an expression of pairwise preference in a~label ranking~\cite{Hllermeier2010}. Unlike the previously mentioned decomposition technique, the pairwise method is adjusted for paired inter-label dependencies. Contrary to the BR approach, during this type of decomposition a~substantially larger number of base classifiers are generated and need to be built. In general, the transformed datasets are less imbalanced than those obtained as a~result of \textit{one-vs-rest} transformation. Moreover, the models created with the base classifiers obtained using this method tend to be simpler than those based on \textit{one-vs-rest} classifiers\cite{Furnkranz2002}.

The primary aim of this study was to compare soft-confusion-matrix approach (SCM) and Bayes metaclassifier (BMC) algorithm under the framework of multi-label classification. Specifically, both techniques were tested for BR and LPW decomposition transformations. Briefly, SCM and BMC algorithms are overlays that can be placed on the top of any base classifier. Moreover, they both derive from the RRC classifier idea introduced by Woloszynski~\cite{Woloszynski2011}. While the SCM and BMC algorithms are constructed using similar ideas, to the best of our knowledge, they have not been compared directly thus far. Therefore, we searched for potential differences between these two approaches.

The concept of the \textit{soft confusion} matrix was first introduced in a~study aimed at improving the classification quality of systems that recognize hand gestures~\cite{Kurzynski2015,Trajdos2016}. In that study, the SCM-based system was used due to its ability to utilize soft class-assignment. Moreover, the system’s potential to improve the response of base classifiers was considered an important argument to use this solution. Also, the \textit{soft confusion} matrix-based approach was employed under a~multi-label classification framework~\cite{Trajdos2015}, to improve the quality of binary relevance classifiers. While the study confirmed validity of this approach, it also demonstrated its sensitivity in the case of unbalanced class distribution in a~binary problem. The algorithm was also used in research on LPW decomposition~\cite{trajdos2017correction,trajdos2017correctionWeight}. Those experiments also showed that the use of the model contributed to a~substantial improvement of the outcome of the committee built using the \textit{one-vs-one} approach.

The concept of Bayes metaclassifier was first introduced in~\cite{MM:Majak2017}. The authors of that study presented in detail and validated the concept of BMC algorithm. The study demonstrated that the upper bound of the BMC improvement over base classifier was a~Bayes error. BMC performance was shown to be directly related to the classification quality of a~base classifier. Moreover, the experiments involving benchmark datasets showed that BMC significantly improved base classifier classification, especially whenever the chosen base classifier was far from optimum. Additionally, analysis of the results from each cross-validation phase demonstrated that BMC contributed to a~decrease in the classification variance when compared with its basic counterpart results. BMC was shown to perform better with balanced datasets. However, if the \textit{a’priori} probabilities of the BMC design are used to solve imbalanced problems, its decision boundary is moved towards a~majority class. Since BMC provides probabilistic interpretation for any base classifier response, this method was also used to address sequential classification problems, and was shown to be a~useful tool for constructing multi-classifier systems, especially during classifier fusion~\cite{MM:MBC:2016, MM:Majak:2016}.

This paper is organized as follows. Section~\ref{sec:Methods} provides formal notation used throughout the article, and introduces the SCM correction algorithm and Bayes metaclassifier. Section~\ref{sec:ExpSetup} contains a~description of the experimental setup. In section~\ref{sec:ResultsAndDiscussion}, the results of the study are presented and discussed, and final conclusions are presented in the~\ref{sec:Conclusions} section.

\section{Methods}\label{sec:Methods}
\subsection{Preliminaries}\label{sec:Methods:prelim}

In this section, a~single-label classification is introduced. The scope is reduced to only binary classification because only binary classifiers are considered in this paper. As stated in the Introduction, in the single-label classification approach, a~$d-\mathrm{dimensional}$ object $\vec{x}=\ptVecX{x}{}{d}\in\mathbb{X}=\mathbb{R}^{d}$ is assigned a~class $m \in\mathbb{M}$, where $\mathbb{M}=\left\{0,1\right\}$ is an output space (a set of available classes). The single-label, binary classifier $\psi: \mathbb{X}\mapsto \mathbb{M}$ is an approximation of an unknown mapping $f: \mathbb{X}\mapsto\mathbb{M}$ which assigns the classes to the instances.  The classification methods analyzed in this paper follow the statistical classification framework. Hence, a~feature vector $\vec{x}$ and its label $m$ are assumed to be realisations of random variables ${\vec{\textbf{X}}}$ and ${\vec{\textbf{M}}}$, respectively. The random variables follow the joint probability distribution $P(\vec{\textbf{X}},\vec{\textbf{M}})$. Given the loss function $l:\mathbb{M}\times\mathbb{M}\mapsto \mathbb{R}^{+}\cup \{0\}$, assessing the similarity of the objects in the output space, the optimal prediction $\psi^{*}(\vec{x})$ for the object $x$ can be calculated as follows:
\begin{align}
 \psi^{*}(\vec{x}) &= \argmin_{\vec{k}\in\mathbb{M}}\sum_{\vec{m}\in\mathbb{M}}l(\vec{k},\vec{m})P(\vec{\textbf{M}}=\vec{m}| \vec{\textbf{X}}=\vec{x}).
\end{align}
If the loss function $l$ is the zero-one loss, the optimal decision is made using the maximum \textit{a posteriori} rule:
\begin{align}
 \psi^{*}(\vec{x}) &=\argmax_{\vec{k}\in\mathbb{M}}P(\mathbb{M}=k|\mathbb{X}=x),
\end{align}
where $P(\mathbb{M}=k|\mathbb{X}=x)$ is the conditional probability that the object $x$ belongs to class $k$.

In this paper, the so-called soft output of the classifier $\nu: \mathbb{X}\mapsto \left [0,1\right ]^{2}$ is also defined. The soft output vector $\nu$ contains values   proportional to the conditional probabilities. Consequently, the following conditions need to be satisfied:
\begin{align}
&\nu_{i} \approx P(\mathbb{M}=i|\mathbb{X}=x),\\
&\nu_{i}(\vec{x}) \in \left [ 0 , 1 \right ], \\
&\sum_{i=0}^{1}\nu_{i}(\vec{x}) = 1.
\end{align}

In this study, a~classifier $\psi$ is built in a~supervised learning procedure using the training set $\mathcal{T}$ containing $|\mathcal{T}|$ pairs of feature vectors $\vec{x}$ and corresponding labels $\vec{m}$:
\begin{equation}\label{eq:trainSet} 
\mathcal{T}=\left\{(\vec{x}^{(1)},m^{(1)}), (\vec{x}^{(2)},m^{(2)}), \ldots ,(\vec{x}^{(|\mathcal{V}|)},m^{(|\mathcal{T}|)})\right\},
\end{equation}
where $\vec{x}^{(k)} \in \mathbb{X}$ and $m^{(k)} \in \mathbb{M}$. To evaluate the classifier the validation set $\mathcal{V}$ was used as well. 

\subsection{Multi-label Classification}\label{sec:Methods:MultiLabel}

Under the Multi-label formalism, an object $x$ is assigned to a~set of labels indicated by a~binary vector of length $L$: $\vec{y}=\ptVecX{y}{}{L}\in\mathbb{Y}=\{0,1\}^{L}$, where $L$ denotes the number of labels. Each element of the binary vector corresponds to a~single label. If for some object $x$ the element of the vector $y_{i}$ is set to $1$ ($0$), this means that the label associated with $i\mathrm{-th}$ position is relevant (irrelevant) to object $x$. Relevant labels are assigned to instances by an unknown mapping $g:\mathbb{X}\mapsto\mathbb{Y}$. A~multi-label classifier $h: \mathbb{X}\mapsto\mathbb{Y}$ is an approximation of the unknown mapping.

The classification process often consists of two steps. During the first step, the classifier produces soft outputs $\omega:\mathbb{X}\mapsto\mathbb{R}^{L}$. Then the classifier's outcome is generated using the thresholding procedure:
\begin{align}\label{eq:thresh}
 h(\vec{x}) = \left[ \ivBrack{\omega_{1}(\vec{x})\geq \Theta_{1}}, \ivBrack{\omega_{2}(\vec{x})\geq \Theta_{2}}, \cdots, \ivBrack{\omega_{L}(\vec{x})\geq \Theta_{L}}  \right], 
\end{align}
where $\ivBrack{\cdot}$ is the Iverson bracket\cite{Knuth1992} and $\Theta_{i}$ is a~label-specific threshold that may be found using various thresholding strategies~\cite{Ioannou2010,Yang2001}.

As mentioned above, as a~result of BR transformation, a~separate binary classifier is obtained for each label. Hence, the BR ensemble consists of $L$ binary classifiers:
\begin{align}
  \mathcal{C}_{\mathrm{BR}} =&  \left\{ \psi^{(1)}, \psi^{(2)},\cdots, \psi^{(L)} \right\}.
\end{align}
The output of the multi-label classifier is obtained as follows:
\begin{align}
 h_{l}(\vec{x})&= \psi^{(l)}(\vec{x}).
\end{align}

The label-pairwise (LPW) transformation produces multi-label classifier $h$, using an ensemble of binary classifiers $\Psi$, and then, a~single binary classifier is assigned to each pair of labels:
\begin{align}
\mathcal{C}_{\mathrm{LPW}}&=\left\{\psi^{(i,j)}| i,j \in \{1,2,\cdots, L\}, i<j\right\},\\
\psi^{(i,j)}(\vec{x}) &\in \{i,j\},\\
\left| \mathcal{C}_{\mathrm{LPW}} \right| &= \frac{L(L-1)}{2}.
\end{align}

The soft output of the LPW ensemble is obtained by combining the outcomes of the base classifiers:
\begin{align}
 \omega_{l}(\vec{x}) = \frac{1}{L-1}\sum_{\substack{i,j\in{1,2,\cdots,L}\\i<j}}\ivBrack{\psi^{(i,j)}(\vec{x})=l}.
\end{align}
The soft output is then converted into a~binary response using a~thresholding procedure\eqref{eq:thresh}.

\subsection{Randomized Reference Classifier}\label{subsect:RRC}

In the hereby presented approaches, the behaviour of a~base classifier $\psi$ was modeled using a~stochastic classifier defined by a~probability distribution over the set of labels $\mathbb{M}$. In this study, the randomized reference classifier (RRC) proposed by Woloszynski and Kurzynski~\cite{Woloszynski2011} was used. The RRC is a~hypothetical classifier that allows a~randomised model of a~given deterministic classifier to be built.

We assumed that for a~given instance $\vec{x}$, the randomised classifier $\psi^{(R)} $ generates a~vector of class supports  $\left[ \nu_{1}(\vec{x}), \nu_{2}(\vec{x})\right]$ being observed values of random variables $\left[ \Delta_{1}(\vec{x}), \Delta_{2}(\vec{x})\right]$. The chosen probability distribution of random variables needs to satisfy the following conditions:
\begin{align}
\label{MK_PT:delta_c1}
\Delta_{1}(\vec{x}), \, \Delta_{2}(\vec{x}) &\in (0,1), \\
\label{MK_PT:delta_c2}
\Delta_{1}(\vec{x})+\Delta_{2}(\vec{x})&=1, \\
\label{MK_PT:delta_c3}
\mathbf{E}\left[\Delta_{i}(\vec{x}) \right] &= \nu_{i}(\vec{x}),\ i \in \{0,1 \},
\end{align}
where $\mathbf{E}$ is the expected value operator. Conditions (\ref{MK_PT:delta_c1})  and (\ref{MK_PT:delta_c2})  follow from the normalisation properties of class supports, whereas condition (\ref{MK_PT:delta_c3}) provides the equivalence of the randomized model $\psi^{(R)}$ and base classifier $\psi$. Based on the latter condition, the RRC can be used to provide a~randomised model of any classifier that returns a~vector of class-specific supports $\nu(\vec{x})$. 

The probability of classifying an object $\vec{x}$ into the class $i$ using the RRC can be calculated from the following formula:
\begin{equation}   \label{MK_PT:wzor5}
P(\mathbf{\Psi}=m|\vec{\textbf{X}}=\vec{x})=Pr\left[\Delta_{m}(\vec{x})> \Delta_{\{0,1 \} \setminus m}(\vec{x})\right],
\end{equation}
where $Pr\left[\Delta_{m}(\vec{x})> \Delta_{\{0,1 \} \setminus m}(\vec{x})\right]$ is the probability that the value obtained by the realisation of random variable $\Delta_{m}$ is greater than the realisation of random variable $\Delta_{\{0,1 \} \setminus m}$.

The key step in the modeling process presented above is selection of the probability distributions for random variables $\Delta_{i}(\vec{x})\;  i \in \{0,1\}$ that satisfy the conditions~\eqref{MK_PT:delta_c1}-\eqref{MK_PT:delta_c3}. In this study, in line with the recommendations given in~\cite{Woloszynski2011}, the beta~distribution with parameters  $\lambda_{i}(\vec{x}),\mu_{i}(\vec{x}),\;  i \in \{0,1\}$ was applied. The parameters were chosen based on the following set of equations:
\begin{equation}\label{eq:betaParamsEq} 
\left\{ \begin{array}{rcl}
 \dfrac{\lambda_{i}(\vec{x})}{\lambda_{i}(\vec{x})+\mu_{i}(\vec{x})}&=&\nu_{i}(\vec{x}),\\ \\
 \lambda_{i}(\vec{x})+\mu_{i}(\vec{x})&=&2.
 \end{array} \right.
\end{equation}

The rationale for the choice of beta distribution based on the theory of order statistics can be found in~\cite{Woloszynski2011} along with a~detailed description of the estimation parameters. For the beta~distribution, the following formula for probability (\ref{MK_PT:wzor5}) was obtained:
\begin{align} 
\label{MK_PT:wzor6} P^{(R)}(\mathbf{\Psi}=m|\vec{\textbf{X}}=\vec{x})  &= \int_0^1 b(u,\lambda_{m}(\vec{x}), \mu_{m}(\vec{x})) B(u,\lambda_{j}(\vec{x})), \mu_{j}(\vec{x})\ du,\ j \neq m,
\end{align}
where $B(\dot)$ is a~beta cumulative distribution function and $b(\dot)$ is a~beta probability density function. 
It needs to be stressed that no validation set is required to calculate the probabilities~\eqref{MK_PT:wzor6}, as knowledge of the correct classification of object $\vec{x}$ is not a~must. 
The MATLAB implementation of the RRC classifier is freely available at~\footnote{\href{http://www.mathworks.com/matlabcentral/fileexchange/28391-a-probabilistic-model-of-classifier-competence}{http://www.mathworks.com/matlabcentral/fileexchange/28391-a-probabilistic-model-of-classifier-competence}}. Implementation for WEKA~is also available~\footnote{\href{https://github.com/ptrajdos/rrcBasedClassifiers/tree/develop}{https://github.com/ptrajdos/rrcBasedClassifiers/tree/develop}}.

\subsection{Soft-confusion Matrix Classifier}\label{sec:Methods:SCM}

The proposed correction method is based on an assessment of the probability of classifying an object $\vec{x}$ into the class $s \in \mathbb{M}$  using the binary classifier $\psi$. It also provides an extension of the Bayesian model in which the object’s description $\vec{x}$ and its true label $m \in \mathbb{M}$ are realizations of random variables  $\vec{\textbf{X}}$ and $\textbf{M}$, respectively.  In the SCM approach, classifier $\psi$ predicts randomly based on the probabilities $P(\mathbf{\Psi}(\vec{x})=s)=P(s|\vec{x})$~\cite{Berger1985}. Hence, the outcome of the classification $s$ is a~realization of the random variable $\mathbf{\Psi}(\vec{x})$.

According to the extended Bayesian model, the posterior probability $P(m|\vec{x})$ of label $m$ can be defined as:
\begin{align}\label{eq:postProb1}
 P(m|\vec{x}) &= \sum_{s \in \mathcal{M}} P(s|\vec{x}) P(m|s,\vec{x}). 
\end{align}
where $P(m|s,\vec{x})$ denotes the probability that an object $\vec{x}$ belongs to the class $m$ given that $\mathbf{\Psi}(\vec{x})=s$.

Unfortunately, the assumption that base classifier assigns labels in a~stochastic way is rather impractical, since most real-life classifiers are deterministic. This issue was addressed by implementation of deterministic binary classifiers in which their statistical properties were modelled using the RRC procedure, as described in section~\ref{subsect:RRC}.

\subsubsection{Confusion Matrix}\label{sec:Methods:SCM:confMatrix}

During the inference phase, the probability $P(m|s,\vec{x}), s \in \mathbb{M}$ was estimated using a~local, soft confusion matrix. An example of such a~matrix for a~binary classification task is given in Table~\ref{MK_PT:confmatrix}. The rows of the matrix correspond to the ground-truth classes, whereas the columns match the outcome of the classifier. The confusion matrix is considered soft because the decision regions of the random classifier are expressed in terms of fuzzy set formalism~\cite{Zadeh1965}. Thus, the membership function of a~point $\vec{x}$ is proportional to the probability of assigning $\vec{x}$ to a~given class using the randomized model of the classifier. 

Based on subsets of the validation set that contain object belonging to class~$m$, the fuzzy decision region of $\psi$ and the neighborhood of $\vec{z}$ are defined according to the formulas:
\begin{align}
\label{eq:fvs}
  \mathcal{V}_{s} &= \left\{  (\vec{x}^{(k)},s^{(k)}, 1): (\vec{x}^{(k)},s^{(k)}) \in \mathcal{V}, s^{(k)}=s  \right\},\\
\label{eq:fds}
  {\mathcal{D}}_{s} &= \left\{  (\vec{x}^{(k)},s^{(k)} , \mu_{\mathcal{D}{s}}(\vec{x}^{(k)}  ) ): (\vec{x}^{(k)},s^{(k)}) \in \mathcal{V}\right\},\\
\label{eq:fns}
 \mathcal{N}(\vec{z}) &= \left\{  (\vec{x^{(k)}},s^{(k)} ,\mu_{\mathcal{N}(\vec{z})}(\vec{x^{(k)}}) ):(\vec{x}^{(k)},s^{(k)}) \in \mathcal{V} \right\},
\end{align}
where each triplet $(\vec{x}^{(k)},s^{(k)}, \zeta)$ defines the fuzzy membership value $\zeta$ of instance $(\vec{x}^{(k)},s^{(k)})$, and $\mu_{\mathcal{D}{s}}(x)=P^{(R)}(s|\vec{x})$ indicates the fuzzy decision region of the stochastic classifier. Additionally, $\mu_{\mathcal{N}(\vec{z})}(\vec{x})$ denotes the fuzzy neighbourhood of the instance $\vec{z}$. The membership function of the neighbourhood is defined using the Gaussian potential function:
\begin{equation}   \label{MK_PT:mu}
 \mu_{\mathcal{N}(\vec{z})}(\vec{x^{(k)}})=\exp({-\beta\delta(\vec{z},\vec{x^{(k)}})^2}),
\end{equation}
where $\beta\in \mathbb{R}_{+}$ and $\delta(\vec{z},\vec{x^{(k)}})$ is a~distance  function between two vectors from the input space $\mathbb{X}$.

The fuzzy sets defined above can be then employed to approximate the entries of the local confusion  matrix:
\begin{align}\label{eq:uFCM}
\hat{\varepsilon}_{m,s}(\vec{z}) &= \frac{|\mathcal{V}_{s} \cap {\mathcal{D}}_{m} \cap \mathcal{N}(\vec{z})|}{|\mathcal{N}(\vec{z})|},
\end{align}
where $|.|$ is the cardinality of a~fuzzy set~\cite{Dhar2013}. Finally, the approximation of $P(s|m,\vec{x})$ is calculated as follows:
\begin{equation} 
\label{pt:postApprox}
P(m|s,\vec{x})  \approx \frac{\hat{\varepsilon}_{m,s}(\vec{z})}{\sum_{u \in \mathbb{M}}\hat{\varepsilon}_{u,s}(\vec{z})}.
\end{equation}


\begin{table}[tb]
\centering\normalsize
\caption{The confusion matrix for a~binary classification problem.\label{MK_PT:confmatrix}}
{\begin{tabular}{cc|cc}
& & \multicolumn{2}{c}{estimated}\\
& &  $s=1$ & $s=2$\\
\hline
\multirow{2}{*}{true}& $m=1$& $\varepsilon_{1,1}$&$\varepsilon_{1,2}$\\
& $m=2$& $\varepsilon_{2,1}$&$\varepsilon_{2,2}$\\
\end{tabular}
}
\end{table}

\subsection{Bayes metaclassifier--BMC}\label{sec:Methods:MB}

First, the probabilistic model of classification will be introduced. The probability distribution of  $(\textbf{X}, \textbf{M})$ is determined based on \emph{a~priori} class probabilities $p_m=P(\textbf{M}=m)$ and class-conditional density functions $f(x|m)=f_m(x)$.

The Bayes metaclassifier (BMC) $\psi^{BMC}$, represents the probabilistic generalization of any base classifier (\ref{eq:bmc1}) which has the form of the Bayes scheme built over the classifier $\psi$. Thus, in $\psi^{BMC}$ approach, the decision is based on the maximum \textit{a~posteriori} probability rule~\cite{MM:Majak2017, MM:MBC:2016}: 
\begin{equation} 
\label{eq:bmc1}
\begin{aligned}
\psi^{BMC}(\psi(x)=s)=\argmax_{k \in \mathbb{M}} \{p(k|\psi=s)\}.
\end{aligned}
\end{equation}

\textit{A~posteriori} probabilities $p(i|s) \equiv P(\textbf{M}=i|\psi(x)=s),  i \in \mathbb{M}$ derive from the Bayes rule:
\begin{equation}   
\label{eq:bmc2}
P(\textbf{M}=i|\psi=s) = \frac{p_i \; p(s|i)}{\sum_j p_j \; p(s|j) },
\end{equation}
where probability $p(s|i) \equiv P(\psi(x)=s| \mathbf{M}=i)$ denotes class-dependent probability of erroneous (if $s \neq i$) or correct (if $s \equiv i$) classification of an object~$\vec{x}$ 
by the base classifier $\psi$.

When the base classifier  $\psi$ is placed in a~probabilistic frame defined by the BMC $\psi^{BMC}$ , a~common probabilistic interpretation of any response of base classifiers is obtained, regardless of the design paradigm.

The key element in the  BMC scheme (\ref{eq:bmc1}) and (\ref{eq:bmc2}) is the calculation of probabilities $P(\psi(x)=s|\mathbf{M}=i)$ at point~$\vec{x}$, i.e. the class-dependent probabilities of correct classification and misclassification with base classifiers. Usually, such probabilities for a~base deterministic classifier would be either $0$ for misclassification or $1$ for correct classification of a~given~$\vec{x}$. However, in this paper, an alternative method for approximating these probabilities was proposed, based on the original concept of a~randomized reference classifier (RRC).

The RRC $\psi^{RRC}(x)$ is a~stochastic classifier defined by a~probability distribution chosen in such a~way, that RRC acts, on average,
as an modeled base classifier. Under such assumption, the class-dependent probabilities of correct classification $P_c(j|x)$ and 
misclassification $P_e(j|x)$ can be calculated as an equivalent to the modeled base classifier:
\begin{equation}    
\label{wzor4a}
P(\psi(x)=s|i) \approx P(\psi^{RRC}(x)=s|i).
\end{equation}

In the computational procedure, the probabilities $P(\psi(x)=s|i) \approx P(\psi^{RRC}(x)=s|i)$ (denoting that an objects $\vec{x}$ belongs to class $i$ given that $\psi(x)=s$) are calculated for each validation point included in the validation set). As these values are only known for discrete points from $\mathcal{V}$, to enable dynamic calculation of any new object $\vec{x}$ during classification, a~neighborhood function is needed to describe how the probabilities at validation points affect the new $\vec{x}$. For the BMC algorithm, Gaussian potential function is used as a~neighborhood function:
\begin{equation}
\label{eq:gaussianPotentialFunction}
\begin{aligned}
P_c^{RRC}(j|x) = \frac{\sum_{x^{(k)} \in \mathcal{V}, j^{(k)}=j} P_c(j|x)\cdot \exp(-\beta\delta(\vec{x},\vec{x^{(k)}})^2)}{\sum_{x^{(k)} \in \mathcal{V}, j^{(k)}=j} \exp(-\beta\delta(\vec{x},\vec{x^{(k)}})^2)}, \\ \\
P_e^{RRC}(j|x) = \frac{\sum_{x^{(k)} \in \mathcal{V}, j^{(k)} \neq j} P_e(j|x)\cdot \exp(-\beta\delta(\vec{x},\vec{x^{(k)}})^2)}{\sum_{x^{(k)} \in \mathcal{V}, j^{(k)} \neq j} \exp(-\beta\delta(\vec{x},\vec{x^{(k)}})^2)},
\end{aligned}
\end{equation}
\noindent where $\beta$ value in equation (\ref{eq:gaussianPotentialFunction}) is a~scaling factor which should be adjusted independently to classification problem. Similarly, \textit{a~priori} probabilities ($p_l, l \in \mathbb{M}$) introduced in (\ref{eq:bmc2}) are estimated using the validation set $\mathcal{V}$.

\section{Experimental Setup}\label{sec:ExpSetup}
The study included two scenarios using BR transformation and LPW approach, respectively. Regardless the scenario, the following methods were compared:
\begin{enumerate}
 \item Unmodified base classifiers,
 \item Base classifiers combined with the Bayes metaclassifier approach,
 \item Base classifiers combined with the SCM approach.
\end{enumerate}

The following single-label classifiers were used during the study:
\begin{itemize}
 \item J48 (C4.5) classifier~\cite{Quinlan1993},
 \item SVM classifier with radial kernel~\cite{Cortes1995,CC01a},
 \item Naive Bayes classifier~\cite{Hand2001},
 \item Nearest Neighbour classifier~\cite{Cover1967}.
\end{itemize}

All the experimental code was implemented using WEKA~\cite{Hall2009}. The base classifiers were also obtained from this framework. During the study, the parameters of the J48 algorithm were set to its defaults. For the naive Bayes classifier,
the kernel estimator with the Gaussian kernel was used to calculate the probabilities. The parameters of the SVM classifier ($C \in \{.001,1,2,\ldots, 10 \}$, $\gamma~\in \{.001,1,2,\ldots, 5\}$) were tuned using grid search and threefold cross-validation. Also the number of the nearest neighbors was tuned using the threefold cross-validation. The number of neighbors was selected from the following values $K \in \{1,3,5,\ldots, 11 \}$. The Euclidean distance function was employed to choose the nearest neighbours. The $\fOne$ criterion calculated for the minority class was used as the quality criterion for the tuning procedures. Other parameters of the base classifiers were set to their defaults.

The Bayes metaclassifier and the SCM-based algorithms were implemented in JAVA. The source code for the algorithms is available online~\footnote{\href{https://github.com/ptrajdos/rrcBasedClassifiers/tree/develop}{https://github.com/ptrajdos/rrcBasedClassifiers/tree/develop}}. 

The size of the neighborhood, expressed as $\beta$ coefficient, was chosen using a~threefold cross-validation procedure and the grid search technique. The search space was defined as follows:
\begin{align*}
 \left\{ \beta = 2+0.9\cdot i, \quad i \in\{0,1,\ldots,10\} \right\}.
\end{align*}
The conversion of soft outputs into binary responses was done using S-Cut algorithm~\cite{Ioannou2010} with the number of cross-validation folds set at three. The thresholds and the size of the committee were chosen in a~way that provided the best value of the $F_1$ criterion~\cite{Rijsbergen1979}. 

To evaluate the proposed methods, the following multi-label classification quality criteria were used~\cite{Luaces2012}:
\begin{itemize}
 \item Hamming loss,
 \item Zero-one loss,
 \item Example based $\fdr$, $\fnr$, $\fOne$,
 \item Macro-averaged $\fdr$, $\fnr$, $\fOne$,
 \item Micro-averaged $\fdr$, $\fnr$, $\fOne$,
\end{itemize}
In line with the recommendations of~\cite{demsar2006} and~\cite{garcia2008extension}, statistical significance of the results was verified using the two-step procedure. The first step was the Friedman test~\cite{Friedman1940} conducted for each quality criterion separately. Since multiple criteria were employed, the familywise errors (\FWER{}) should be controlled~\cite{benjamini2001control}. Thus, the Holm’s~\cite{holm1979} procedure was used to control the \FWER{} of the Friedman tests. Whenever the Friedman test demonstrated a~significant difference within the group of classifiers, the pairwise comparisons were conducted with the Wilcoxon signed-rank test~\cite{wilcoxon1945,demsar2006}. To control the \FWER{} of the Wilcoxon tests, the Holm approach was employed~\cite{holm1979}. The level of statistical significance for all tests was set at $\alpha=0.05$.

Table~\ref{tab:BenchmarkSetsCharacteristics} presents the collection of the benchmark sets that were used during the experimental evaluation of the proposed algorithms. The table is organized as follows. The first column contains the names of the datasets. The names under which the datasets are registered in the repositories were used. The second column contains the numbers of the datasets preceded by the number of the table. Further columns contain the set-specific characteristics of the benchmark sets: 
\begin{itemize}
 \item The number of instances in the dataset ($|S|$),
 \item Dimensionality of the input space ($d$),
 \item The number of labels ($L$),
 \item Average number of labels for a~single instance ($\mathrm{LC}$),
 \item The number of unique label combinations ($\mathrm{LU}$),
 \item Average between-labels imbalance ratio ($\mathrm{IR}$).
\end{itemize}

The datasets are available online~\footnote{\href{https://github.com/ptrajdos/MLResults/blob/master/data/ThesisBenchmark.zip}{https://github.com/ptrajdos/MLResults/blob/master/data/ThesisBenchmark.zip}}. During the preprocessing stage, the datasets underwent a~few transformations. First, all nominal attributes, except binary attributes, were converted into a~set of binary variables. This approach is one of the simplest methods to replace nominal variables with binary variables~\cite{Tian2005}. The transformation is necessary whenever the SVM-based or distance-based algorithms are employed~\cite{Tian2005}. The features were also normalized to have zero mean value and zero unit variance.

In this study, some datasets that follow multi-instance-multi-label (\MIML{}) (datasets: \ref{set:azotobacter}, \ref{set:caenorhabditis}, \ref{set:drosophila}, \ref{set:geobacter}, \ref{set:haloarcula}, \ref{set:pyrococcus}, \ref{set:saccharomyces},\ref{set:mimlImg}) framework~\cite{Zhou2007} were employed. In these datasets, each object consists of a~bag of instances tagged with a~set of labels. To tackle these data, we followed the recommendation of~\cite{Zhou2012miml}, transforming the set into single-instance multi-label data. The multi-target regression sets (datasets:~\ref{set:flare}, \ref{set:water}) were also harnessed. The datasets were converted into multi-label data, using a~simple thresholding procedure. Specifically, when the value of output variable for a~given object was greater than zero, the corresponding label was set to be relevant to this object. The number of labels in the stackex\_chess (\ref{set:stackex}) dataset was reduced to 15, to reduce the computational burden. Moreover, features were selected using a~correlation-based approach before the learning phase~\cite{Hall1999}.

Both training and testing datasets were extracted using tenfold cross-validation. The validation set was essentially the same as the training set, but the base-classifier responses were obtained using twofold cross-validation.

{
\setlength\tabcolsep{2.0pt}%
\def\arraystretch{0.7}%
\footnotesize%
\begin{table}
 \centering\normalsize
 \caption{The characteristics of the benchmark sets}\label{tab:BenchmarkSetsCharacteristics}
 \begin{tabular}{c|Nc|cccccc}

 Name&\multicolumn{1}{c}{No}&Src&$|\mathcal{S}|$&$d$&$L$&$\mathrm{LC}$&$\mathrm{UC}$&$\mathrm{IR}$\\
 \hline
arts1&	\label{set:arts}&\cite{ueda2003parametric}	&7484&	1733&	26&	1.65&	599&	94.74\\
azotobacter\_vinelandii&\label{set:azotobacter}&\cite{Wu2014}	&407&	20&	13&	1.47&	31&	2.23\\
birds&	\label{set:birds}&\cite{Briggs2012}	&645&	260&	19&	1.01&	133&	5.41\\
caenorhabditis\_elegans&	\label{set:caenorhabditis}&\cite{Wu2014}	&2512&	20&	21&	2.42&	65&	2.35\\
drosophila\_melanogaster&	\label{set:drosophila}&\cite{Wu2014}	&2605&	20&	22&	2.66&	63&	1.74\\
emotions&	\label{set:emotions}&\cite{Trohidis2008}	&593&	72&	6&	1.87&	27&	1.48\\
enron&	\label{set:enron}&\cite{enron_Berkeley}	&1702&	1001&	53&	3.38&	753&	73.95\\
flags&	\label{set:flags}&\cite{UCI2013}	&194&	43&	7&	3.39&	54&	2.25\\
flare&	\label{set:flare}&\cite{flare1973}	&1066&	27&	3&	0.21&	7&	14.15\\
genbase&	\label{set:genbase}&\cite{Diplaris2005}	&662&	1186&	27&	1.25&	32&	37.31\\
geobacter-sulfurreducens&	\label{set:geobacter}&\cite{Wu2014}	&379&	20&	11&	1.26&	28&	2.75\\
haloarcula\_marismortui&	\label{set:haloarcula}&\cite{Wu2014}	&304&	20&	13&	1.60&	29&	2.42\\
human&	\label{set:human3160}&\cite{Xu2013}	&3106&	440&	14&	1.19&	85&	15.29\\
IMDB&	\label{set:IMDB}&\cite{kong2011multi}	&3042&	1001&	28&	1.99&	587&	24.61\\
LLOG&	\label{set:LLOG}&\cite{Read2011}	&1460&	1004&	75&	1.18&	304&	39.27\\
medical&	\label{set:medical}&\cite{Pestian2007}	&978&	1449&	45&	1.25&	94&	89.50\\
mimlImg&	\label{set:mimlImg}&\cite{Zhou2007}	&2000&	135&	5&	1.24&	20&	1.19\\
ohsumed&	\label{set:ohsumed}&\cite{Hersh1994}	&13929&	1002&	23&	1.66&	1147&	7.87\\
plant&	\label{set:plant978}&\cite{Xu2013}	&978&	440&	12&	1.08&	32&	6.69\\
pyrococcus\_furiosus&	\label{set:pyrococcus}&\cite{Wu2014}	&425&	20&	18&	2.14&	45&	2.42\\
saccharomyces\_cerevisiae&	\label{set:saccharomyces}&\cite{Wu2014}	&3509&	20&	27&	2.27&	109&	2.08\\
scene&	\label{set:scene}&\cite{Boutell2004}	&2407&	294&	6&	1.07&	15&	1.25\\
simpleHC&	\label{set:simpleHC}&\cite{Tomas2014}	&3000&	30&	10&	1.90&	294&	1.14\\
simpleHS&	\label{set:simpleHS}&\cite{Tomas2014}	&3000&	30&	10&	2.31&	364&	2.62\\
SLASHDOT&	\label{set:SLASHDOT}&\cite{Read2011}	&3782&	1079&	22&	1.18&	156&	17.69\\
stackex\_chess&	\label{set:stackex}&\cite{charte2015quinta}	&1675&	585&	15&	1.14&	139&	4.74\\
tmc2007&	\label{set:tmc2007}&\cite{srivastava2005discovering}	&2857&	500&	22&	2.22&	396&	17.15\\
water-quality&	\label{set:water}&\cite{Deroski2000}	&1060&	16&	14&	5.07&	825&	1.77\\
yeast&	\label{set:yeast}&\cite{Elisseeff2001}	&2417&	103&	14&	4.24&	198&	7.20\\
 \end{tabular}
\end{table}
}

\section{Results and Discussion}\label{sec:ResultsAndDiscussion}

To compare multiple algorithms on multiple benchmark sets, the average ranks approach~\cite{demsar2006} was used. In this approach, the winning algorithm achieves rank equal ’1’, the second achieves rank equal ’2’, etc. In the case of ties, the ranks for algorithms that achieve the same results are averaged. The average ranks are visualized on radar plots. Visualization properties of the radar plots are similar to the properties of parallel coordinates plots. In other words, radar plots can be interpreted as parallel coordinates plots drawn in polar coordinate systems~\cite{Saary2008}. In the plots, the data are visualized in such way that the lowest ranks are closer to the center of the graph. The radar plots illustrating the results of the present experiment are shown in \figurename~\ref{fig:BR} and~\ref{fig:LPW}.

The numerical results are presented in Tables~\ref{table:BR-J48} -- \ref{table:LPW-KNN}. Each table has the same structure. The first row contains numbers assigned to algorithms in section~\ref{sec:ExpSetup}. Then the table is divided into eleven sections, each corresponding to a~single evaluation criterion. The first row of each section is the name of the analyzed quality criterion. The second row contains p-value for the Friedman test, whereas the average ranks for the algorithms are shown in the third row. Further rows contain p-values for the pairwise Wilcoxon tests. The p-value equal to 0.000 depicts p-values lower than $10^{-3}$ and p-value equal to $1.000$ corresponds to p-values greater than $0.999$.

Complete results of the study are available online~\footnote{\href{https://github.com/ptrajdos/MLResults/tree/master/BMAndSCM}{https://github.com/ptrajdos/MLResults/tree/master/BMAndSCM}}.

\subsection{Binary Relevance}\label{sec:ResultsAndDiscussion:BR}
The results related to the binary relevance transformation are presented in figure~\ref{fig:BR} and tables~\ref{table:BR-J48} -- \ref{table:BR-KNN}. The results seem to be partially inconclusive. While no statistically significant between-method differences in all base classifiers and quality criteria were found on the Friedman tests, the significant method-related differences were observed between some classifiers on the post-hoc Wilcoxon tests. A~few trends can be observed in the analyzed data. First, regardless the base classifier, both methods had the same average ranks for macro-averaged FDR, macro and micro-averaged $F_1$ measures. Moreover, for macro-averaged $F_1$ criterion, the post-hoc test demonstrated that for all base classifiers except KNN, the Bayes metaclassifier outperformed the reference approach. Also for macro-averaged FDR, the Bayes metaclassifier significantly outperformed the reference method for three out of four classifiers (except the naive Bayes classifier). These findings imply that the Bayes metaclassifier provided better classification quality for rare labels. Regarding the macro-averaged measures, based on the average ranks, the SCM-based classifier seems to be slightly less conservative than the Bayes metaclassifier. However, the differences between the two approaches were statistically significant only in the case of SVM base classifier.

No consistent conclusions can be formulated from the example-based quality criteria. The order of investigated classifiers (according to their average ranks) varied depending on the base classifier. Also, the results of the post-hoc tests were inconsistent. For these quality criteria, no significant differences were found between the investigated classifiers. Furthermore, no evident trend was observed in the results for the micro-averaged criteria.

\begin{table}[ht]
\centering\normalsize
\caption{Binary Relevance transformation. Wilcoxon test for J48 base classifiers -- p-values for paired comparisons of investigated methods.\label{table:BR-J48}}
\begin{tabular}{c|ccc|ccc|ccc|ccc}
  & 1 & 2 & 3 & 1 & 2 & 3 & 1 & 2 & 3 & 1 & 2 & 3 \\ 
  \hline
Nam.&\multicolumn{3}{c|}{Hamming}&\multicolumn{3}{c|}{Zero-One}&\multicolumn{3}{c|}{ExFDR}&\multicolumn{3}{c}{ExFNR}\\
Frd.&\multicolumn{3}{c|}{1.000e+00}&\multicolumn{3}{c|}{1.000e+00}&\multicolumn{3}{c|}{1.000e+00}&\multicolumn{3}{c}{9.677e-01}\\
 \cmidrule(lr){2-4}\cmidrule(lr){5-7}\cmidrule(lr){8-10}\cmidrule(lr){11-13}
Rank & 1.983 & 1.879 & 2.138 & 2.190 & 1.983 & 1.828 & 2.086 & 1.741 & 2.172 & 2.069 & 2.207 & 1.724 \\ 
  1 &  & 0.741 & 0.138 &  & 0.548 & 0.785 &  & 0.723 & 1.000 &  & 0.933 & 0.395 \\ 
  2 &  &  & 0.087 &  &  & 0.785 &  &  & 1.000 &  &  & 0.331 \\ 
   \hline
Nam.&\multicolumn{3}{c|}{ExF1}&\multicolumn{3}{c|}{MaFDR}&\multicolumn{3}{c|}{MaFNR}&\multicolumn{3}{c}{MaF1}\\
Frd.&\multicolumn{3}{c|}{1.000e+00}&\multicolumn{3}{c|}{4.658e-01}&\multicolumn{3}{c|}{2.183e-01}&\multicolumn{3}{c}{4.559e-01}\\
 \cmidrule(lr){2-4}\cmidrule(lr){5-7}\cmidrule(lr){8-10}\cmidrule(lr){11-13}
Rank & 2.052 & 1.776 & 2.172 & 2.293 & 1.672 & 2.034 & 2.241 & 2.172 & 1.586 & 2.362 & 1.879 & 1.759 \\ 
  1 &  & 0.999 & 1.000 &  & 0.040 & 0.336 &  & 0.182 & 0.073 &  & 0.007 & 0.073 \\ 
  2 &  &  & 1.000 &  &  & 0.287 &  &  & 0.096 &  &  & 0.205 \\ 
   \hline
Nam.&\multicolumn{3}{c|}{MiFDR}&\multicolumn{3}{c|}{MiFNR}&\multicolumn{3}{c|}{MiF1}&\multicolumn{3}{c}{}\\
Frd.&\multicolumn{3}{c|}{1.000e+00}&\multicolumn{3}{c|}{1.277e-01}&\multicolumn{3}{c|}{4.658e-01}&\multicolumn{3}{c}{}\\
 \cmidrule(lr){2-4}\cmidrule(lr){5-7}\cmidrule(lr){8-10}
Rank & 2.052 & 1.914 & 2.034 & 2.241 & 2.207 & 1.552 & 2.328 & 1.948 & 1.724 &  &  &  \\ 
  1 &  & 1.000 & 1.000 &  & 0.464 & 0.036 &  & 0.094 & 0.101 &  &  &  \\ 
  2 &  &  & 1.000 &  &  & 0.028 &  &  & 0.230 &  &  &  \\ 
  \end{tabular}
\end{table}

\begin{table}[ht]
\centering\normalsize
\caption{Binary Relevance transformation. Wilcoxon test for SVM base classifiers -- p-values for paired comparisons of investigated methods.\label{table:BR-SVM}}
\begin{tabular}{c|ccc|ccc|ccc|ccc}
  & 1 & 2 & 3 & 1 & 2 & 3 & 1 & 2 & 3 & 1 & 2 & 3 \\ 
  \hline
Nam.&\multicolumn{3}{c|}{Hamming}&\multicolumn{3}{c|}{Zero-One}&\multicolumn{3}{c|}{ExFDR}&\multicolumn{3}{c}{ExFNR}\\
Frd.&\multicolumn{3}{c|}{9.579e-01}&\multicolumn{3}{c|}{1.000e+00}&\multicolumn{3}{c|}{1.000e+00}&\multicolumn{3}{c}{1.000e+00}\\
 \cmidrule(lr){2-4}\cmidrule(lr){5-7}\cmidrule(lr){8-10}\cmidrule(lr){11-13}
Rank & 1.741 & 2.017 & 2.241 & 1.914 & 2.017 & 2.069 & 1.845 & 2.086 & 2.069 & 1.983 & 2.155 & 1.862 \\ 
  1 &  & 0.108 & 0.005 &  & 1.000 & 1.000 &  & 0.999 & 0.999 &  & 0.776 & 0.693 \\ 
  2 &  &  & 0.005 &  &  & 1.000 &  &  & 0.999 &  &  & 0.570 \\ 
   \hline
Nam.&\multicolumn{3}{c|}{ExF1}&\multicolumn{3}{c|}{MaFDR}&\multicolumn{3}{c|}{MaFNR}&\multicolumn{3}{c}{MaF1}\\
Frd.&\multicolumn{3}{c|}{1.000e+00}&\multicolumn{3}{c|}{5.110e-01}&\multicolumn{3}{c|}{5.611e-02}&\multicolumn{3}{c}{1.217e-01}\\
 \cmidrule(lr){2-4}\cmidrule(lr){5-7}\cmidrule(lr){8-10}\cmidrule(lr){11-13}
Rank & 1.948 & 2.017 & 2.034 & 2.259 & 1.672 & 2.069 & 2.397 & 2.052 & 1.552 & 2.397 & 1.983 & 1.621 \\ 
  1 &  & 1.000 & 1.000 &  & 0.020 & 0.405 &  & 0.017 & 0.017 &  & 0.004 & 0.023 \\ 
  2 &  &  & 1.000 &  &  & 0.024 &  &  & 0.029 &  &  & 0.156 \\ 
   \hline
Nam.&\multicolumn{3}{c|}{MiFDR}&\multicolumn{3}{c|}{MiFNR}&\multicolumn{3}{c|}{MiF1}&\multicolumn{3}{c}{}\\
Frd.&\multicolumn{3}{c|}{2.130e-01}&\multicolumn{3}{c|}{1.692e-01}&\multicolumn{3}{c|}{1.000e+00}&\multicolumn{3}{c}{}\\
 \cmidrule(lr){2-4}\cmidrule(lr){5-7}\cmidrule(lr){8-10}
Rank & 1.603 & 2.121 & 2.276 & 2.293 & 2.121 & 1.586 & 2.224 & 2.017 & 1.759 &  &  &  \\ 
  1 &  & 0.007 & 0.030 &  & 0.052 & 0.022 &  & 0.181 & 0.181 &  &  &  \\ 
  2 &  &  & 0.096 &  &  & 0.026 &  &  & 0.442 &  &  &  \\ 
  \end{tabular}
\end{table}

\begin{table}[ht]
\centering\normalsize
\caption{Binary Relevance transformation. Wilcoxon test for Naive Bayes base classifiers -- p-values for paired comparisons of investigated methods.\label{table:BR-NB}}
\begin{tabular}{c|ccc|ccc|ccc|ccc}
  & 1 & 2 & 3 & 1 & 2 & 3 & 1 & 2 & 3 & 1 & 2 & 3 \\ 
  \hline
Nam.&\multicolumn{3}{c|}{Hamming}&\multicolumn{3}{c|}{Zero-One}&\multicolumn{3}{c|}{ExFDR}&\multicolumn{3}{c}{ExFNR}\\
Frd.&\multicolumn{3}{c|}{1.000e+00}&\multicolumn{3}{c|}{1.000e+00}&\multicolumn{3}{c|}{1.000e+00}&\multicolumn{3}{c}{1.000e+00}\\
 \cmidrule(lr){2-4}\cmidrule(lr){5-7}\cmidrule(lr){8-10}\cmidrule(lr){11-13}
Rank & 2.052 & 1.845 & 2.103 & 2.224 & 1.845 & 1.931 & 2.121 & 1.707 & 2.172 & 1.879 & 2.017 & 2.103 \\ 
  1 &  & 0.741 & 0.299 &  & 0.999 & 1.000 &  & 0.426 & 0.966 &  & 1.000 & 1.000 \\ 
  2 &  &  & 0.217 &  &  & 1.000 &  &  & 0.859 &  &  & 1.000 \\ 
   \hline
Nam.&\multicolumn{3}{c|}{ExF1}&\multicolumn{3}{c|}{MaFDR}&\multicolumn{3}{c|}{MaFNR}&\multicolumn{3}{c}{MaF1}\\
Frd.&\multicolumn{3}{c|}{1.000e+00}&\multicolumn{3}{c|}{1.000e+00}&\multicolumn{3}{c|}{1.000e+00}&\multicolumn{3}{c}{1.000e+00}\\
 \cmidrule(lr){2-4}\cmidrule(lr){5-7}\cmidrule(lr){8-10}\cmidrule(lr){11-13}
Rank & 2.155 & 1.741 & 2.103 & 2.259 & 1.776 & 1.966 & 2.086 & 2.052 & 1.862 & 2.259 & 1.914 & 1.828 \\ 
  1 &  & 0.809 & 0.809 &  & 0.084 & 0.530 &  & 0.600 & 0.467 &  & 0.036 & 0.311 \\ 
  2 &  &  & 0.448 &  &  & 0.733 &  &  & 0.600 &  &  & 0.565 \\ 
   \hline
Nam.&\multicolumn{3}{c|}{MiFDR}&\multicolumn{3}{c|}{MiFNR}&\multicolumn{3}{c|}{MiF1}&\multicolumn{3}{c}{}\\
Frd.&\multicolumn{3}{c|}{1.000e+00}&\multicolumn{3}{c|}{1.000e+00}&\multicolumn{3}{c|}{2.928e-01}&\multicolumn{3}{c}{}\\
 \cmidrule(lr){2-4}\cmidrule(lr){5-7}\cmidrule(lr){8-10}
Rank & 1.983 & 2.017 & 2.000 & 2.086 & 2.017 & 1.897 & 2.397 & 1.879 & 1.724 &  &  &  \\ 
  1 &  & 1.000 & 1.000 &  & 0.570 & 0.570 &  & 0.021 & 0.395 &  &  &  \\ 
  2 &  &  & 1.000 &  &  & 0.570 &  &  & 0.442 &  &  &  \\ 
  \end{tabular}
\end{table}

\begin{table}[ht]
\centering\normalsize
\caption{Binary Relevance transformation. Wilcoxon test for KNN base classifiers -- p-values for paired comparisons of investigated methods.\label{table:BR-KNN}}
\begin{tabular}{c|ccc|ccc|ccc|ccc}
  & 1 & 2 & 3 & 1 & 2 & 3 & 1 & 2 & 3 & 1 & 2 & 3 \\ 
  \hline
Nam.&\multicolumn{3}{c|}{Hamming}&\multicolumn{3}{c|}{Zero-One}&\multicolumn{3}{c|}{ExFDR}&\multicolumn{3}{c}{ExFNR}\\
Frd.&\multicolumn{3}{c|}{3.827e-01}&\multicolumn{3}{c|}{8.242e-02}&\multicolumn{3}{c|}{1.000e+00}&\multicolumn{3}{c}{1.000e+00}\\
 \cmidrule(lr){2-4}\cmidrule(lr){5-7}\cmidrule(lr){8-10}\cmidrule(lr){11-13}
Rank & 2.207 & 1.621 & 2.172 & 2.466 & 1.810 & 1.724 & 2.207 & 1.759 & 2.034 & 2.121 & 2.017 & 1.862 \\ 
  1 &  & 0.002 & 0.137 &  & 0.003 & 0.008 &  & 0.039 & 0.324 &  & 0.241 & 0.197 \\ 
  2 &  &  & 0.024 &  &  & 0.115 &  &  & 0.782 &  &  & 0.241 \\ 
   \hline
Nam.&\multicolumn{3}{c|}{ExF1}&\multicolumn{3}{c|}{MaFDR}&\multicolumn{3}{c|}{MaFNR}&\multicolumn{3}{c}{MaF1}\\
Frd.&\multicolumn{3}{c|}{6.670e-01}&\multicolumn{3}{c|}{6.176e-01}&\multicolumn{3}{c|}{1.000e+00}&\multicolumn{3}{c}{1.000e+00}\\
 \cmidrule(lr){2-4}\cmidrule(lr){5-7}\cmidrule(lr){8-10}\cmidrule(lr){11-13}
Rank & 2.310 & 1.897 & 1.793 & 2.276 & 1.690 & 2.034 & 2.017 & 2.190 & 1.793 & 2.241 & 1.931 & 1.828 \\ 
  1 &  & 0.100 & 0.324 &  & 0.044 & 0.798 &  & 0.431 & 0.338 &  & 0.209 & 1.000 \\ 
  2 &  &  & 0.594 &  &  & 0.338 &  &  & 0.263 &  &  & 1.000 \\ 
   \hline
Nam.&\multicolumn{3}{c|}{MiFDR}&\multicolumn{3}{c|}{MiFNR}&\multicolumn{3}{c|}{MiF1}&\multicolumn{3}{c}{}\\
Frd.&\multicolumn{3}{c|}{3.827e-01}&\multicolumn{3}{c|}{1.000e+00}&\multicolumn{3}{c|}{6.670e-01}&\multicolumn{3}{c}{}\\
 \cmidrule(lr){2-4}\cmidrule(lr){5-7}\cmidrule(lr){8-10}
Rank & 2.379 & 1.759 & 1.862 & 1.948 & 2.224 & 1.828 & 2.310 & 1.931 & 1.759 &  &  &  \\ 
  1 &  & 0.050 & 0.785 &  & 0.361 & 0.324 &  & 0.330 & 1.000 &  &  &  \\ 
  2 &  &  & 0.785 &  &  & 0.152 &  &  & 1.000 &  &  &  \\ 
  \end{tabular}
\end{table}

\begin{figure}[tb]
 \centering\normalsize
  \subfloat[J48\label{fig:BR:J48}]{\includegraphics[width = 0.49\textwidth]{\ptFiguresDirectory{radarBR-J48}}}
  \subfloat[SVM\label{fig:BR:SVM}]{\includegraphics[width = 0.49\textwidth]{\ptFiguresDirectory{radarBR-SVM}}}\\
  \subfloat[NB\label{fig:BR:NB}]{\includegraphics[width = 0.49\textwidth]{\ptFiguresDirectory{radarBR-NB}}}
  \subfloat[KNN\label{fig:BR:KNN}]{\includegraphics[width = 0.49\textwidth]{\ptFiguresDirectory{radarBR-KNN}}}
  \caption{Average ranks of for Binary Relevance approach.\label{fig:BR}}
\end{figure}

\subsection{Label Pairwise}\label{sec:ResultsAndDiscussion:LPW}
The results for the label-pairwise transformation are shown in figure~\ref{fig:LPW} and tables~\ref{table:LPW-J48} -- \ref{table:LPW-KNN}. The results seem to be consistent; whenever the p-value for the Friedman test was significant, at least one significant between-algorithm difference was found on the post-hoc Wilcoxon test.

The statistical analysis demonstrated that regardless the base classifier, the Bayes metaclassifier and the classifier based on the soft confusion matrix were more conservative than the reference method, i.e. provided significantly better results
in terms of the FDR (precision) criterion and significantly worse outcomes in terms of the FNR (recall) criterion. Consequently, the analyzed methods performed significantly better in terms of the $F_1$ measure. In other words, the proposed methods were more conservative as they identified fewer instances as relevant, but a~larger proportion of the identified instances turned out to be truly relevant for the instance under classification. The results are consistent across all example-based, micro and macro-averaged criteria. This means that the results were better from the perspective of the whole label vector and each label separately, whether common or rare in the dataset. Moreover, the analyzed algorithms significantly outperformed the reference method in terms of the Hamming loss and the zero-one loss. The improvement in terms of the zero-one loss seems to be particularly important as this is the most strict quality criterion that can be used for quality assessment of multi-label classifiers. That is to say, the criterion assigns the loss equal to 1 if only a~single label is misclassified. The results of our present study clearly show that unlike for the BR-based approach, the analyzed algorithms contributed to a~substantial improvement of classification quality for the label-pairwise transformation. A~reason behind the evident quality improvement for the label-pairwise transformation might be the fact that the datasets obtained by the LPW transformation were markedly less imbalanced than those obtained by the BR transformation. As a~result, the neighborhood of a~point was predominated by points belonging to the majority class. If the neighborhood is imbalanced, statistical properties of the base classifiers cannot be estimated appropriately.

No statistically significant differences were found between the Bayes metaclassifier and the classifier based on the confusion matrix. This is not so surprising since both analyzed algorithms are based on similar principles, namely, they both use a~validation set to provide a~probabilistic interpretation of a~base classifier.

\begin{table}[ht]
\centering\normalsize
\caption{Pairwise transformation. Wilcoxon test for J48 base classifiers -- p-values for paired comparisons of investigated methods.\label{table:LPW-J48}}
\begin{tabular}{c|ccc|ccc|ccc|ccc}
  & 1 & 2 & 3 & 1 & 2 & 3 & 1 & 2 & 3 & 1 & 2 & 3 \\ 
  \hline
Nam.&\multicolumn{3}{c|}{Hamming}&\multicolumn{3}{c|}{Zero-One}&\multicolumn{3}{c|}{ExFDR}&\multicolumn{3}{c}{ExFNR}\\
Frd.&\multicolumn{3}{c|}{5.916e-09}&\multicolumn{3}{c|}{7.088e-09}&\multicolumn{3}{c|}{5.498e-09}&\multicolumn{3}{c}{4.744e-08}\\
 \cmidrule(lr){2-4}\cmidrule(lr){5-7}\cmidrule(lr){8-10}\cmidrule(lr){11-13}
Rank & 3.000 & 1.500 & 1.500 & 2.946 & 1.375 & 1.679 & 3.000 & 1.607 & 1.393 & 1.071 & 2.536 & 2.393 \\ 
  1 &  & 0.000 & 0.000 &  & 0.000 & 0.000 &  & 0.000 & 0.000 &  & 0.000 & 0.000 \\ 
  2 &  &  & 0.884 &  &  & 0.243 &  &  & 0.479 &  &  & 0.126 \\ 
   \hline
Nam.&\multicolumn{3}{c|}{ExF1}&\multicolumn{3}{c|}{MaFDR}&\multicolumn{3}{c|}{MaFNR}&\multicolumn{3}{c}{MaF1}\\
Frd.&\multicolumn{3}{c|}{8.157e-05}&\multicolumn{3}{c|}{5.916e-09}&\multicolumn{3}{c|}{2.801e-08}&\multicolumn{3}{c}{2.392e-03}\\
 \cmidrule(lr){2-4}\cmidrule(lr){5-7}\cmidrule(lr){8-10}\cmidrule(lr){11-13}
Rank & 2.679 & 1.786 & 1.536 & 3.000 & 1.571 & 1.429 & 1.071 & 2.643 & 2.286 & 2.536 & 1.714 & 1.750 \\ 
  1 &  & 0.000 & 0.000 &  & 0.000 & 0.000 &  & 0.000 & 0.000 &  & 0.000 & 0.000 \\ 
  2 &  &  & 0.264 &  &  & 0.508 &  &  & 0.245 &  &  & 0.779 \\ 
   \hline
Nam.&\multicolumn{3}{c|}{MiFDR}&\multicolumn{3}{c|}{MiFNR}&\multicolumn{3}{c|}{MiF1}&\multicolumn{3}{c}{}\\
Frd.&\multicolumn{3}{c|}{5.916e-09}&\multicolumn{3}{c|}{4.710e-09}&\multicolumn{3}{c|}{3.969e-06}&\multicolumn{3}{c}{}\\
 \cmidrule(lr){2-4}\cmidrule(lr){5-7}\cmidrule(lr){8-10}
Rank & 3.000 & 1.571 & 1.429 & 1.000 & 2.643 & 2.357 & 2.786 & 1.750 & 1.464 &  &  &  \\ 
  1 &  & 0.000 & 0.000 &  & 0.000 & 0.000 &  & 0.000 & 0.000 &  &  &  \\ 
  2 &  &  & 0.779 &  &  & 0.236 &  &  & 0.255 &  &  &  \\ 
  \end{tabular}
\end{table}

\begin{table}[ht]
\centering\normalsize
\caption{Pairwise transformation. Wilcoxon test for SVM base classifiers -- p-values for paired comparisons of investigated methods.\label{table:LPW-SVM}}
\begin{tabular}{c|ccc|ccc|ccc|ccc}
  & 1 & 2 & 3 & 1 & 2 & 3 & 1 & 2 & 3 & 1 & 2 & 3 \\ 
  \hline
Nam.&\multicolumn{3}{c|}{Hamming}&\multicolumn{3}{c|}{Zero-One}&\multicolumn{3}{c|}{ExFDR}&\multicolumn{3}{c}{ExFNR}\\
Frd.&\multicolumn{3}{c|}{4.282e-09}&\multicolumn{3}{c|}{9.299e-09}&\multicolumn{3}{c|}{6.585e-09}&\multicolumn{3}{c}{6.601e-08}\\
 \cmidrule(lr){2-4}\cmidrule(lr){5-7}\cmidrule(lr){8-10}\cmidrule(lr){11-13}
Rank & 3.000 & 1.643 & 1.357 & 2.964 & 1.518 & 1.518 & 3.000 & 1.464 & 1.536 & 1.071 & 2.464 & 2.464 \\ 
  1 &  & 0.000 & 0.000 &  & 0.000 & 0.000 &  & 0.000 & 0.000 &  & 0.000 & 0.000 \\ 
  2 &  &  & 0.438 &  &  & 0.576 &  &  & 0.814 &  &  & 0.646 \\ 
   \hline
Nam.&\multicolumn{3}{c|}{ExF1}&\multicolumn{3}{c|}{MaFDR}&\multicolumn{3}{c|}{MaFNR}&\multicolumn{3}{c}{MaF1}\\
Frd.&\multicolumn{3}{c|}{6.601e-08}&\multicolumn{3}{c|}{6.601e-08}&\multicolumn{3}{c|}{1.727e-07}&\multicolumn{3}{c}{9.899e-08}\\
 \cmidrule(lr){2-4}\cmidrule(lr){5-7}\cmidrule(lr){8-10}\cmidrule(lr){11-13}
Rank & 2.929 & 1.571 & 1.500 & 2.929 & 1.536 & 1.536 & 1.143 & 2.357 & 2.500 & 2.893 & 1.500 & 1.607 \\ 
  1 &  & 0.000 & 0.000 &  & 0.000 & 0.000 &  & 0.000 & 0.000 &  & 0.000 & 0.000 \\ 
  2 &  &  & 1.000 &  &  & 0.662 &  &  & 0.438 &  &  & 0.991 \\ 
   \hline
Nam.&\multicolumn{3}{c|}{MiFDR}&\multicolumn{3}{c|}{MiFNR}&\multicolumn{3}{c|}{MiF1}&\multicolumn{3}{c}{}\\
Frd.&\multicolumn{3}{c|}{3.415e-09}&\multicolumn{3}{c|}{6.585e-09}&\multicolumn{3}{c|}{9.631e-09}&\multicolumn{3}{c}{}\\
 \cmidrule(lr){2-4}\cmidrule(lr){5-7}\cmidrule(lr){8-10}
Rank & 3.000 & 1.679 & 1.321 & 1.000 & 2.464 & 2.536 & 2.964 & 1.679 & 1.357 &  &  &  \\ 
  1 &  & 0.000 & 0.000 &  & 0.000 & 0.000 &  & 0.000 & 0.000 &  &  &  \\ 
  2 &  &  & 0.056 &  &  & 0.630 &  &  & 0.412 &  &  &  \\ 
  \end{tabular}
\end{table}
\begin{table}[ht]
\centering\normalsize
\caption{Pairwise transformation. Wilcoxon test for Naive Bayes base classifiers -- p-values for paired comparisons of investigated methods.\label{table:LPW-NB}}
\begin{tabular}{c|ccc|ccc|ccc|ccc}
  & 1 & 2 & 3 & 1 & 2 & 3 & 1 & 2 & 3 & 1 & 2 & 3 \\ 
  \hline
Nam.&\multicolumn{3}{c|}{Hamming}&\multicolumn{3}{c|}{Zero-One}&\multicolumn{3}{c|}{ExFDR}&\multicolumn{3}{c}{ExFNR}\\
Frd.&\multicolumn{3}{c|}{4.710e-09}&\multicolumn{3}{c|}{7.641e-09}&\multicolumn{3}{c|}{4.710e-09}&\multicolumn{3}{c}{4.949e-08}\\
 \cmidrule(lr){2-4}\cmidrule(lr){5-7}\cmidrule(lr){8-10}\cmidrule(lr){11-13}
Rank & 3.000 & 1.643 & 1.357 & 2.964 & 1.607 & 1.429 & 3.000 & 1.643 & 1.357 & 1.107 & 2.393 & 2.500 \\ 
  1 &  & 0.000 & 0.000 &  & 0.000 & 0.000 &  & 0.000 & 0.000 &  & 0.000 & 0.000 \\ 
  2 &  &  & 0.508 &  &  & 0.782 &  &  & 0.386 &  &  & 0.920 \\ 
   \hline
Nam.&\multicolumn{3}{c|}{ExF1}&\multicolumn{3}{c|}{MaFDR}&\multicolumn{3}{c|}{MaFNR}&\multicolumn{3}{c}{MaF1}\\
Frd.&\multicolumn{3}{c|}{1.323e-08}&\multicolumn{3}{c|}{4.710e-09}&\multicolumn{3}{c|}{3.558e-08}&\multicolumn{3}{c}{3.558e-08}\\
 \cmidrule(lr){2-4}\cmidrule(lr){5-7}\cmidrule(lr){8-10}\cmidrule(lr){11-13}
Rank & 2.964 & 1.607 & 1.429 & 3.000 & 1.643 & 1.357 & 1.071 & 2.464 & 2.464 & 2.929 & 1.607 & 1.464 \\ 
  1 &  & 0.000 & 0.000 &  & 0.000 & 0.000 &  & 0.000 & 0.000 &  & 0.000 & 0.000 \\ 
  2 &  &  & 0.567 &  &  & 0.245 &  &  & 0.955 &  &  & 0.218 \\ 
   \hline
Nam.&\multicolumn{3}{c|}{MiFDR}&\multicolumn{3}{c|}{MiFNR}&\multicolumn{3}{c|}{MiF1}&\multicolumn{3}{c}{}\\
Frd.&\multicolumn{3}{c|}{4.710e-09}&\multicolumn{3}{c|}{1.323e-08}&\multicolumn{3}{c|}{4.710e-09}&\multicolumn{3}{c}{}\\
 \cmidrule(lr){2-4}\cmidrule(lr){5-7}\cmidrule(lr){8-10}
Rank & 3.000 & 1.643 & 1.357 & 1.036 & 2.536 & 2.429 & 3.000 & 1.607 & 1.393 &  &  &  \\ 
  1 &  & 0.000 & 0.000 &  & 0.000 & 0.000 &  & 0.000 & 0.000 &  &  &  \\ 
  2 &  &  & 0.316 &  &  & 0.630 &  &  & 0.630 &  &  &  \\ 
  \end{tabular}
\end{table}
\begin{table}[ht]
\centering\normalsize
\caption{Pairwise transformation. Wilcoxon test for KNN base classifiers -- p-values for paired comparisons of investigated methods.\label{table:LPW-KNN}}
\begin{tabular}{c|ccc|ccc|ccc|ccc}
  & 1 & 2 & 3 & 1 & 2 & 3 & 1 & 2 & 3 & 1 & 2 & 3 \\ 
  \hline
Nam.&\multicolumn{3}{c|}{Hamming}&\multicolumn{3}{c|}{Zero-One}&\multicolumn{3}{c|}{ExFDR}&\multicolumn{3}{c}{ExFNR}\\
Frd.&\multicolumn{3}{c|}{7.937e-08}&\multicolumn{3}{c|}{6.152e-08}&\multicolumn{3}{c|}{9.242e-08}&\multicolumn{3}{c}{2.475e-07}\\
 \cmidrule(lr){2-4}\cmidrule(lr){5-7}\cmidrule(lr){8-10}\cmidrule(lr){11-13}
Rank & 2.929 & 1.643 & 1.429 & 2.911 & 1.393 & 1.696 & 2.929 & 1.500 & 1.571 & 1.107 & 2.500 & 2.393 \\ 
  1 &  & 0.000 & 0.000 &  & 0.000 & 0.000 &  & 0.000 & 0.000 &  & 0.000 & 0.000 \\ 
  2 &  &  & 0.508 &  &  & 0.319 &  &  & 0.745 &  &  & 0.339 \\ 
   \hline
Nam.&\multicolumn{3}{c|}{ExF1}&\multicolumn{3}{c|}{MaFDR}&\multicolumn{3}{c|}{MaFNR}&\multicolumn{3}{c}{MaF1}\\
Frd.&\multicolumn{3}{c|}{2.087e-06}&\multicolumn{3}{c|}{8.048e-09}&\multicolumn{3}{c|}{6.910e-07}&\multicolumn{3}{c}{2.145e-05}\\
 \cmidrule(lr){2-4}\cmidrule(lr){5-7}\cmidrule(lr){8-10}\cmidrule(lr){11-13}
Rank & 2.821 & 1.607 & 1.571 & 3.000 & 1.536 & 1.464 & 1.143 & 2.357 & 2.500 & 2.714 & 1.679 & 1.607 \\ 
  1 &  & 0.000 & 0.000 &  & 0.000 & 0.000 &  & 0.000 & 0.000 &  & 0.000 & 0.000 \\ 
  2 &  &  & 0.728 &  &  & 0.236 &  &  & 0.955 &  &  & 0.374 \\ 
   \hline
Nam.&\multicolumn{3}{c|}{MiFDR}&\multicolumn{3}{c|}{MiFNR}&\multicolumn{3}{c|}{MiF1}&\multicolumn{3}{c}{}\\
Frd.&\multicolumn{3}{c|}{6.954e-08}&\multicolumn{3}{c|}{9.242e-08}&\multicolumn{3}{c|}{9.570e-06}&\multicolumn{3}{c}{}\\
 \cmidrule(lr){2-4}\cmidrule(lr){5-7}\cmidrule(lr){8-10}
Rank & 2.929 & 1.679 & 1.393 & 1.071 & 2.500 & 2.429 & 2.750 & 1.750 & 1.500 &  &  &  \\ 
  1 &  & 0.000 & 0.000 &  & 0.000 & 0.000 &  & 0.000 & 0.000 &  &  &  \\ 
  2 &  &  & 0.126 &  &  & 0.614 &  &  & 0.236 &  &  &  \\ 
  \end{tabular}
\end{table}

\begin{figure}[tb]
 \centering\normalsize
  \subfloat[J48\label{fig:LPW:J48}]{\includegraphics[width = 0.49\textwidth]{\ptFiguresDirectory{radarLPW-J48}}}
  \subfloat[SVM\label{fig:LPW:SVM}]{\includegraphics[width = 0.49\textwidth]{\ptFiguresDirectory{radarLPW-SVM}}}\\
  \subfloat[NB\label{fig:LPW:NB}]{\includegraphics[width = 0.49\textwidth]{\ptFiguresDirectory{radarLPW-NB}}}
  \subfloat[KNN\label{fig:LPW:KNN}]{\includegraphics[width = 0.49\textwidth]{\ptFiguresDirectory{radarLPW-KNN}}}
  \caption{Average ranks of for Label Pairwise approach.\label{fig:LPW}}
\end{figure}

\section{Conclusions}\label{sec:Conclusions}

In this study, we compared Bayes metaclassifier and soft-confusion-matrix-based classifier. The classifiers were compared with each other and with the reference method under the multi-label classification framework. Specifically, the classifiers were compared using problem-transformation approach to multi-label learning. Two most common transformation methods were used: binary relevance and label-pairwise.

The study showed that:
\begin{itemize}
 \item For the binary-relevance transformation, virtually no significant differences existed between the compared algorithms. While the Bayes metaclassifier seems to be slightly better than the reference method, no statistically significant differences were found between this classifier and the SCM-based method.
 \item For the label-pairwise transformation, both the Bayes metaclassifier and the SCM-based classifier significantly outperformed the reference method in terms of all quality criteria. The analyzed algorithms seem to be more conservative than the reference method. No significant differences were found between the Bayes metaclassifier and the SCM-based classifier.
\end{itemize}


\clearpage
\bibliography{bibliography}

\end{document}